# Bipolar possibilistic representations

Salem Benferhat, Didier Dubois, Souhila Kaci and Henri Prade
Institut de Recherche en Informatique de Toulouse (I.R.I.T.)–C.N.R.S.
Université Paul Sabatier, 118 route de Narbonne 31062 TOULOUSE Cedex 4, FRANCE
E-mail:{benferhat, dubois, kaci, prade}@irit.fr

## Abstract

Recently, it has been emphasized that the possibility theory framework allows us to distinguish between i) what is possible because it is not ruled out by the available knowledge, and ii) what is possible for sure. This distinction may be useful when representing knowledge, for modelling values which are not impossible because they are consistent with the available knowledge on the one hand, and values guaranteed to be possible because reported from observations on the other hand. It is also of interest when expressing preferences, to point out values which are positively desired among those which are not rejected. This distinction can be encoded by two types of constraints expressed in terms of necessity measures and in terms of guaranteed possibility functions, which induce a pair of possibility distributions at the semantic level. A consistency condition should ensure that what is claimed to be guaranteed as possible is indeed not impossible. The present paper investigates the representation of this bipolar view, including the case when it is stated by means of conditional measures, or by means of comparative context-dependent constraints. The interest of this bipolar framework, which has been recently stressed for expressing preferences, is also pointed out in the representation of diagnostic knowledge.

## 1 Introduction

The idea of equipping a set of possible worlds, such as the interpretations of a classical logic language, with a complete pre-order is simple and useful for encoding preferences or plausibility orderings. This idea has emerged in the late eighties for providing non-monotonic consequence relations of the type "from $p$, $q$ normally follows" requiring that all the most preferred/plausible/normal models of $p$ be models of $q$ [15]. Interestingly enough, possibility theory [17, 8] has been developed in the meantime, as a framework for modelling states of partial ignorance on the basis of possibility distributions encoding complete pre-orders. Then, possibility measures of events, defined as the maximum of the distribution over the set of models of the considered events, together with the dual measures of necessity, are used for assessing uncertainty. As such, possibility theory relies on a minimal specificity principle which states that the possibility level associated to a possible world should be the largest one which is compatible with the constraints representing the available information. It agrees in particular with the classical view of logical modelling according to which each of piece of information declares some worlds impossible (the others forming a subset of tentatively possible ones). When pieces of information are conjunctively combined, the resulting set of the possible worlds is indeed the largest set compatible with all pieces of information. Possibility theory extends this view by allowing intermediary levels between what is fully possible and what is completely impossible.

More recently, the idea that there exist situations where the information could be bipolar has been advocated in [16], and in [5] for the possibility theory framework. It corresponds to situations where two complete preorders, rather than one, are necessary to represent the whole information conveniently. While a consistency condition should be maintained between them, they behave in opposite ways in order to be complementary. Namely, while one of the two distributions agrees with the classical logical view mentioned above, the other rather corresponds to an idea of close world assumption, where only the information explicitly reported is held for true. The latter view obeys to a maximal specificity principle where only the worlds which are known as existing are considered to be possible. The information is then accumulated disjunctively in the sense that the larger the number of reported cases, the larger the range of worlds which are guaranteed to be possible.

This bipolar framework can be of interest for modelling knowledge in order to distinguish between what is possible because it is just consistent with what is known, and what is possible for sure since it has been observed. For instance, the information about the prices of houses in some area is usually partly made of general rules constraining the possible prices and of examples of reported cases. The bipolar view is also natural when modelling preferences if we have to distinguish between positive desires and preferences reflecting what it is not rejected as unsatisfactory [3]. Another situation which can lead to a bipolar representation is the case where we start with a possibility distribution over a power set of possible worlds that we want to approximate from above and from below by distributions on the referential set, as explained in Section 7.

The paper aims at showing how the possibility theory framework is convenient for handling bipolar representations. The next section explains how a pair of consistent distributions, referred to as lower and upper distributions respectively, should be associated with a treatment respectively in terms of guaranteed possibility functions (also called $\Delta$-functions) and maximal specificity principle, and in terms of (classical) possibility measure $\Pi$ and minimal specificity principle. A joint treatment of $\Delta$ and $\Pi$ functions is provided, applying maximal specificity principle on $\Delta$, and minimal specificity principle on $\Pi$. It is shown that the two principles are the two sides of a unique principle of minimal commitment. Section 3 summarizes logical representation results in terms of $\Delta$-functions-based constraints and of necessity-based constraints. Sections 4, 5 and 6 show how the idea of conditioning, in terms of context-dependent comparative constraints, or in terms of conditional functions, can be applied to the bipolar representation framework. Section 7 provides an illustration of the usefulness of the proposed representation framework in diagnosis modelling.

## 2 Bipolar possibility theory

### 2.1 Constraints on a possibility distribution

The basic representational tool in possibility theory is the notion of a possibility distribution [17], denoted by $\pi$, which encodes a complete preorder on a referential set U of possible worlds or possible interpretations. A degree $\pi(u) \in [0, 1]$ evaluates to what extent $u \in U$ can be the actual state of the world. So the larger $\pi(u)$, the greater the plausibility of $u$, the more normal $u$ is, the more preferred $u$ is, the more satisfactory $u$ is, according to the problem under concern (knowledge or preference modelling).

A possibility distribution, representing the available information, should obey a *principle of minimal specificity*. This principle states that each possible interpretation $u$ should receive the largest possibility degree in agreement with the set of constraints induced by the available information. If the available information states that "the actual state of the world is in a (classical) subset A", this means that any possibility distribution $\pi$ such that $\forall u, \pi(u) \leq A(u)$ (we use the same notation for a subset and its characteristic function) is in accordance with the information. The information $\pi \leq A$ only rules out the values outside $A$ which are asserted as completely impossible (since $A(u) = 0$ iff $u \notin A$). Choosing a particular $\pi$ such that $\pi < A$, for representing the information, would be arbitrarily too precise. Hence choosing the largest possibility distribution $\pi^*$ such that $\pi \leq A$, i.e. $\pi^* = A$, is natural. Clearly, a collection of constraints $\pi \leq A_i$ for $i = 1, \cdots, n$ should be aggregated conjunctively, i.e., $\pi \leq min_i A_i$. The more constraints of this form, the more precise our information about the location of the actual state of the world. It corresponds to the situation in classical logic, where the more formulas, the smaller the set of interpretations which make them true. The full consistency of the constraints is achieved when $\pi^*$ is *normalized*, i.e. $\exists u, \pi^*(u) = 1$, which means that there remains at least one fully possible interpretation.

This approach extends to the case where the $A_i$'s or $A$ are fuzzy subsets. Let $A_{\overline{\alpha}} = \{u : A(u) > \alpha\}$. Note that the $A_{\overline{\alpha}}$'s are nested, i.e. $\alpha \geq \beta \Rightarrow A_{\overline{\alpha}} \subseteq A_{\overline{\beta}}$. Then $\pi \leq A$ means that $\forall \alpha, \forall u \notin A_{\overline{\alpha}}, \pi(u) \leq \alpha$, i.e. the possibility of $u$ is upper-bounded by $\alpha$. This can be translated in terms of the set of constraints $\forall \alpha, \Pi((A_{\overline{\alpha}})^c) \leq \alpha$, where $A^c$ is the complement of $A$ in $U$ and where $\Pi$ is a *possibility measure* defined from $\pi$ by

$$\Pi(A) = sup_{u \in A} \pi(u), \forall A \subseteq U.$$

It can be as well stated in terms of the dual necessity measure $N(A) = 1 - \Pi(A^c)$, i.e.

$$\forall \alpha, N(A_{\overline{\alpha}}) \geq 1 - \alpha, \qquad (1)$$

which expresses that we are certain, at least at level $1 - \alpha$, that the actual world is in $A_{\overline{\alpha}}$.

But there exists a converse type of constraint of the form $\pi \geq A$, which expresses that all the values in $A$ are possible at least at the degree to which these values belong to $A$. It corresponds to a different kind of information of the type "all the values in $A$ are possible". Such a piece of information is natural when reporting on observed data whose feasibility, or possibility of appearance is thus guaranteed. The aggregation of such pieces of information is disjunctive since $\pi \geq A_j$ for $j = 1, \cdots, m$ entails $\pi \geq max_j A_j$ and corresponds to a data accumulation process. The notion of possibility underlying the constraint $\pi \geq A$ is related to a set function, denoted $\Delta$, different from the possibility measure $\Pi$. Indeed $\pi \geq A$ means that $\forall \alpha, \forall u \in A_\alpha, \pi(u) \geq \alpha$, i.e.

$$\forall \alpha, \Delta(A_\alpha) \geq \alpha, \qquad (2)$$

where $A_\alpha = \{u : A(u) \geq \alpha\}$ and $\Delta(A) = inf_{u \in A}\pi(u)$ is the minimum degree of possibility over $A$ and is called the *"guaranteed possibility"* of $A$.

Clearly with constraints of the form $\pi \geq A$, we cannot apply a principle of minimal specificity (which would lead to the state of complete ignorance, $\forall u, \pi(u) = 1$). We should rather use the converse *principle of maximal specificity* which allocates to each value its minimum degree of possibility enforced by the constraint. It corresponds to the idea of focusing only on what has been actually observed. So the information $\pi \geq A$ translates into $\pi_* = A$ applying this principle. In particular $\pi_*(u) = 0$ if $u \notin \{u : A(u) > 0\}$, since then there is no reason to assume that $u$ is (somewhat) possible, as far as it has not been reported.

Note that $\Pi$ is a maxitive set function (increasing in the wide sense with set inclusion), i.e. $\Pi(A \cup B) = max(\Pi(A), \Pi(B)), \forall A, \forall B$, while guaranteed possibility functions are decreasing set functions such that $\Delta(A \cup B) = min(\Delta(A), \Delta(B))$. This means that interpretations covered by $A$ or $B$ are guaranteed to be possible (i.e., because they are observed, feasible, satisfactory, permitted, according to the problems) if and only if both the interpretations in $A$ and those in $B$ are guaranteed to be possible.

### 2.2 The bipolar view

Generally, the two types of information described above may be available, namely the ones of the form $\pi \leq A_1$ and those of the form $\pi \geq A_2$, leading to the obvious *consistency* requirement $\pi_* = A_2 \leq \pi^* = A_1$, i.e.,

$$\forall u, \pi_*(u) \leq \pi^*(u). \qquad (3)$$

This agrees with the idea that what is observed is only a part of what is possible.

Thus a pair of consistent lower and upper possibility distributions $\pi_*$ and $\pi^*$, obeying to (3) can be seen as encoding two types of information of the form respectively $\Delta(A_{*\alpha}) \geq \alpha$ where $\pi_* = A_*$, i.e. the worlds in $A_{*\alpha}$ are guaranteed to be possible at degree $\alpha$ on the one hand, and of the form $N(A_{\overline{\alpha}}^*) \geq 1 - \alpha$ where $\pi^* = A^*$, i.e. it is certain at degree $1 - \alpha$ that the actual world is in $A_{\overline{\alpha}}^*$.

Given a consistent pair of possibility distributions $\pi_*$ and $\pi^*$ ($\pi_* \leq \pi^*$), the uncertainty of an event $A$ is evaluated by the ordered pair $(\Delta(A), \Pi(A))$, with $\Delta(A) \leq \Pi(A)$, since $\Delta$ is defined from $\pi_*$ and $\Pi$ from $\pi^*$. This can be symbolically written $(\Delta, \Pi)(A) = (\Delta(A), \Pi(A))$. Then we have

$$(\Delta, \Pi)(A \cup B) = mM((\Delta, \Pi)(A), (\Delta, \Pi)(B)), \quad (4)$$

where $m$ stands for $min$ and $M$ for $max$, and $mM((a, b), (c, d)) = (min(a, c), max(b, d))$. Introducing by duality

$$(N, \nabla)(A) = (N(A), \nabla(A)) = 1 - (\Delta, \Pi)(A^c) \quad (5)$$
$$= 1 - (\Delta(A^c), \Pi(A^c))$$
$$= (1 - \Pi(A^c), 1 - \Delta(A^c)),$$

where $1 - (a, b) = (1 - b, 1 - a)$, we have similarly

$$(N, \nabla)(A \cap B) = mM((N, \nabla)(A), (N, \nabla)(B)).$$

$\nabla(A)$ is called the *potential necessity* of $A$. Indeed, we have $N(A) \leq \nabla(A)$. Providing that $\pi^*$ is normalized and that $\exists u, \pi_*(u) = 0$ (which is always possible by adding an extra element), the following inequality holds

$$max(N(A), \Delta(A)) \leq min(\Pi(A), \nabla(A)). \quad (6)$$

Up to (6), the four set functions $N, \Delta, \Pi$ and $\nabla$ are unrelated. This strongly contrasts with the probabilistic setting where only one quantity assesses the uncertainty of $A$, since $Prob(A^c) = 1 - Prob(A)$.

In general, while (4) holds for the disjunction, we only have the inequality

$$\Pi(A \cap B) \leq min(\Pi(A), \Pi(B)),$$

for possibility measures, since $A \cap B$ may be impossible while $A \cap B^c$ and thus $A$ can be fully possible, as well as $A^c \cap B$ and thus $B$. A counterpart of this inequality holds for $\Delta$, namely

$$\Delta(A \cap B) \geq max(\Delta(A), \Delta(B)),$$

since the minimum of $\pi_*$ over $A \cap B$ may be greater than the minima over $A$, and over $B$. These two inequalities can be written more compactly as

$$(\Delta, \Pi)(A \cap B) \subseteq Mm((\Delta, \Pi)(A), (\Delta, \Pi)(B)), \quad (7)$$

where $Mm((a, b), (c, d)) = (max(a, c), min(b, d))$ and where $(a, b) \subseteq (c, d)$ means $a \geq c$ and $b \leq d$. The partial order "$\subseteq$" could be understood here as meaning "better informed than". Note that the principles of minimal and maximal specificity jointly applied to upper and lower constraints $\pi^* \leq A_1$ and $\pi_* \geq A_2$, i.e. $(\pi_*, \pi^*) \subseteq (A_2, A_1)$, leading to take $(\pi_*, \pi^*) = (A_2, A_1)$, amounts to a *unique principle of minimal commitment* regarding the available information: do not represent more information than you have. The principle of minimal commitment applied to (7) leads to the equality

$$(\Delta, \Pi)(A \cap B) = Mm((\Delta, \Pi)(A), (\Delta, \Pi)(B)). \quad (8)$$

When (8) holds, $\Delta(A \cap B)$ (resp. $\Pi(A \cap B)$) does not bring further information w.r.t. $\Delta(A)$ and $\Delta(B)$ (resp. $\Pi(A)$ and $\Pi(B)$).

Observe that (8) encompasses the condition $\Pi(A \cap B) = min(\Pi(A), \Pi(B))$, which is a form of independence between $A$ and $B$, called by Nahmias [13], "unrelatedness". The corresponding form of *independence* for $\Delta$ reads, $\Delta(A \cap B) = max(\Delta(A), \Delta(B))$, i.e. if $A$ and $B$ are unrelated, the interpretations in $A$ and $B$ are guaranteed to be possible as far as the interpretations of $A$, or those of $B$, are guaranteed to be possible.

Besides, apart from '$\subseteq$', there exists another natural partial ordering which can be introduced in the bipolar framework, that we shall denote by $\leq$, which means "*less possible than*", and which is defined by $(\Delta, \Pi)(A) \leq (\Delta, \Pi)(B)$ iff $\Delta(A) \leq \Delta(B)$ and $\Pi(A) \leq \Pi(B)$. These two orderings induce a *bilattice* structure in the sense of Ginsberg [11].

## 3　Bipolar possibilistic logic

Lower and upper possibility distributions are not always directly available, but are rather implicitly specified by means of guaranteed possibility-based constraints and necessity-based constraints respectively. These constraints bear on sets of possible worlds which are no longer necessarily nested, as in the case of the $\alpha$-level cuts of the fuzzy sets specifying the distributions in Section 2.1.

In possibilistic logic [6], the information is prioritized in terms of necessity measures. A *possibilistic logic base* is a set of weighted formulas of the form $\Sigma = \{(a_i, \gamma_i) : i = 1, \cdots, n\}$, where $a_i$ is a propositional formula (which corresponds to a set of models $A_i$) and $\gamma_i$ belongs to a totally ordered scale such as $[0, 1]$. $(a_i, \gamma_i)$ means that the necessity degree of $A_i$ is at least equal to $\gamma_i$, i.e. $N(A_i) \geq \gamma_i$.

With a possibilistic logic base $\Sigma$, a unique possibility distribution, denoted by $\pi_\Sigma$, is associated at the semantical level; it is defined by [6]: $\forall u \in U$,

$$\pi^*_\Sigma(u) = \begin{cases} 1 & \text{if } \forall (a_i, \gamma_i) \in \Sigma, u \models a_i \\ 1 - max\{\gamma_i : (a_i, \gamma_i) \in \Sigma \text{ and } u \not\models a_i\} & \text{otherwise,} \end{cases}$$

where $\pi^*_\Sigma$ is in agreement with the minimal specificity principle. Indeed this is the largest possibility distribution whose the associated necessity measure $N$ satisfies $N(A_i) \geq \gamma_i$ for all $i$ such that $(a_i, \gamma_i) \in \Sigma$.

Besides, the inference rule $(a \vee b, \alpha), (\neg a \vee c, \beta) \vdash (b \vee c, min(\alpha, \beta))$ is the basis of the possibilistic inference machinery at the syntactic level [6].

In possibilistic logic each formula $(a_i, \gamma_i)$ expresses that it is certain at level $\gamma_i$ that the actual state of the world is among the models of $a_i$, i.e. that the counter-models of $a_i$ are only possible at most at level $1 - \gamma_i$. Then a possibilistic logic base $\Sigma = \{(a_i, \gamma_i) : i = 1, \cdots, n\}$ is associated with the upper possibility distribution $\pi^*_\Sigma$ defined above.

Another type of information is encoded by constraints of the form $\Delta(B_j) \geq \delta_j$ for asserting that any model of a proposition $b_j$, whose set of models is $B_j$, is a possible candidate for being the actual world at least at level $\delta_j$. Such a set of constraints defines what can be called a lower possibilistic base, of the form $\Xi = \{[b_j, \delta_j] : j = 1, \cdots, m\}$, where $[b_j, \delta_j]$ encodes the inequality $\Delta(B_j) \geq \delta_j$. Applying the maximal specificity principle, leads to derive the following lower distribution with $\Xi$, defined by

**Definition 1**

$$\pi_{*\Xi}(u) = \begin{cases} 0 & \text{if } \not\exists j, u \models b_j \text{ and } [b_j, \delta_j] \in \Xi \\ max\{\delta_j : u \models b_j, [b_j, \delta_j] \in \Xi\} & \text{otherwise.} \end{cases}$$

$\pi_{*\Xi}$ is the smallest distribution whose associated $\Delta$-function satisfies $\forall j, \Delta(B_j) \geq \delta_j$.

Besides, the logical machinery for $\Delta$-weighted formulas is governed by the following cut rule $[a \wedge b, \alpha], [\neg a \wedge c, \beta] \vdash [b \wedge c, min(\alpha, \beta)]$ [5], which is the counterpart of the possibilistic resolution rule, exchanging $\Delta$ and $N$, $\wedge$ and $\vee$.

## 4　Bipolar conditioning

A conditional possibility measure $\Pi(. \mid A)$ has been defined by Hisdal [12] as satisfying the Bayes-like relationship:

$$\Pi(A \cap B) = min(\Pi(B \mid A), \Pi(A)). \quad (9)$$

The principle of minimal specificity leads to define $\Pi(. \mid A)$ as the greatest solution of this equation, namely, if $A \neq \emptyset$:

$$\Pi(B \mid A) = \begin{cases} 1 & \text{if } \Pi(A \cap B) = \Pi(A) \\ \Pi(A \cap B) & \text{if } \Pi(A \cap B^c) > \Pi(A \cap B). \end{cases}$$

$\Pi(B \mid A) = \Pi(B)$ implies $\Pi(A \cap B) = min(\Pi(A), \Pi(B))$, i.e. $A$ and $B$ are unrelated in the sense of Section 2 (not conversely). The dual conditional necessity measure is simply $N(B \mid A) = 1 - \Pi(B^c \mid A)$.

This is an ordinal form of conditioning, adapted to finite settings, with a qualitative (ordinal) possibility scale. The corresponding conditional possibility distribution $\pi(. \mid A)$ is defined by:

$$\pi(u \mid A) = \begin{cases} 1 & \text{if } u \in A \text{ and } \pi(u) = \Pi(A) \\ \pi(u) & u \in A \text{ and } \pi(u) < \Pi(A) \\ 0 & u \notin A. \end{cases}$$

Conditioning can be also defined for guaranteed possibility measures. As for the inference, it works in a reversed way w.r.t. $\Pi$. Namely, conditioning obeys the following equation

$$\Delta(A \cap B) = max(\Delta(B \mid A), \Delta(A)), \quad (10)$$

It means that $A \cap B$ is guaranteed possible either because $A$ is so or because, in the context $A$, $B$ is guaranteed possible.

Since $\Delta(A) = min(\Delta(A \cap B), \Delta(A \cap B^c))$, it expresses either that the minimum of $\pi_*$ over $A$ is reached on $A \cap B$ and that $\Delta(B \mid A)$ can be taken as equal to 0 by virtue of the principle of maximal specificity, or that $\Delta(A) = \Delta(A \cap B^c) < \Delta(A \cap B)$ and then $\Delta(B \mid A)$ should be equal to $\Delta(A \cap B)$. Thus, we have

$$\Delta(B \mid A) = \begin{cases} \Delta(A \cap B) & \text{if } \Delta(A) < \Delta(A \cap B) \\ 0 & \text{if } \Delta(A \cap B^c) \geq \Delta(A \cap B) = \Delta(A). \end{cases}$$

It can be checked that if $\Delta(B \mid A) = 0$ then $A$ and $B$ are unrelated, i.e. $\Delta(A \cap B) = max(\Delta(A), \Delta(B))$.
Interestingly enough, we can define upper and lower possibilistic conditioning in a compact way as obeying the relation

$$(\Delta, \Pi)(A \cap B) = Mm((\Delta, \Pi)(B \mid A), (\Delta, \Pi)(A)). \quad (11)$$

Moreover it can be shown that
$N(B \mid A) > 0$ iff $\Pi(A \cap B) > \Pi(A \cap B^c)$
iff $N(A \Rightarrow B) > N(A \Rightarrow B^c)$, where $\Rightarrow$ denotes material implication. It expresses that $B$ is somewhat certain in context $A$ iff $A \cap B$ is strictly more plausible than $A \cap B^c$. A similar relation holds for guaranteed possibility: $\Delta(B \mid A) > 0$ iff $\Delta(A \cap B) > \Delta(A \cap B^c)$ iff $\nabla(A \Rightarrow B) > \nabla(A \Rightarrow B^c)$.
This points out the fact that the definition of conditioning enforces the following normalization condition:
$min(\Delta(B \mid A), \Delta(B^c \mid A)) = 0$.
This expresses that everything (i.e. $B$ and $B^c$) cannot be guaranteed as being simultaneously possible, and the fact that $B$ is guaranteed as being possible in context $A$ means that we have more guarantees in favor of $B$ than in favor of $B^c$.
Thus one can write $N(B \mid A) > 0$ and $\Delta(B \mid A) > 0$ iff $(\Delta, \Pi)(A \cap B) > (\Delta, \Pi)(A \cap B^c)$
using the second partial ordering introduced in Section 2.

## 5　Strict comparative possibility bases

A strict comparative possibility base $\mathcal{P}$ is a set of constraints of the form "in context $a$, $b$ is more possible than $\neg b$" i.e. in terms of models, $\Pi(A \cap B) > \Pi(A \cap B^c)$, denoted by $a \to b$. It means that $b$ is true in all best models of $a$. This can either express a general rule having exceptions, or the conditional preference of an agent. It encompasses the general case of constraints of the form $\Pi(A) > \Pi(B)$. Indeed $\Pi(A) > \Pi(B)$ is trivially equivalent to $\Pi(A \cap B^c) > \Pi(B)$ and corresponds to the rule $a \vee b \to \neg b$ since $A \cap B^c = (A \cup B) \cap B^c$.

In a qualitative setting, a possibility distribution $\pi$ can be represented by its well-ordered partition $WOP(\pi) = E_1 \cup \cdots \cup E_n$ such that: $E_1 \cup \cdots \cup E_n = U$, $E_i \cap E_j = \emptyset$, and $\forall u, u', \pi(u) > \pi(u')$ iff $u \in E_i, u' \in E_j$ and $i < j$.

Each strict comparative possibility base $\mathcal{P}$ induces a unique upper qualitative possibility distribution $WOP(\pi_\mathcal{P})$ obtained by considering the least specific solution satisfying:

$$\Pi(A_k \cap B_k) > \Pi(A_k \cap B_k^c) \quad (12)$$

for all $a_k \to b_k$ of $\mathcal{P}$.
In [4], an algorithm has been provided to compute $WOP(\pi_\mathcal{P}) = E_1 \cup \cdots \cup E_n$. Here we only recall its basic principle, which consists in putting each interpretation in the lowest possible rank (or highest possibility degree) without violating constraints (12). The only case where we cannot put $u$ in some partition $E_i$, is when $u$ is in the right part of some constraint (where there is a rule $a_k \to b_k$ such that $u \models a_k \wedge \neg b_k$), and none of the interpretations of the left part of this constraint (i.e., $u \models a_k \wedge b_k$) is already classified in some $E_j$ with $j < i$. Therefore $WOP(\pi_\mathcal{P})$ is computed as follows: for each step $i$, we put in $E_i$ all interpretations which are not in the right part of any constraint, then we remove all rules $a_k \to b_k$ such that there exists at least $u$ in $E_i$ such that $u \models a_k \wedge b_k$.

There exists a converse transformation from $\pi$ to $\mathcal{P}$, see [2].

Similarly, a set of $\Delta$-based comparative constraints $\Delta(A_k \cap B_k) > \Delta(A_k \cap B_k^c)$ defines a lower qualitative possibility distribution by applying the maximal specificity principle.
Let $G = \{a_k \leadsto b_k : k = 1, \cdots, n\}$ be a set of rules. Let us denote by $\mathcal{C} = \{\Delta(A_k \cap B_k) > \Delta(A_k \cap B_k^c) : a_k \leadsto b_k \in G\}$ the set of $\Delta$-based comparative constraints induced by $G$. Algorithm 1 provides the most specific qualitative distribution satisfying $G$, denoted by $WOP(\pi_G)$.

The idea of the algorithm consists in assigning to each interpretation the lowest possibility degree. At each step $i$, we put in $S_i$ the interpretations which are not in left part of any $\Delta$-based constraint $\Delta(A_k \cap B_k) > \Delta(A_k \cap B_k^c)$ (otherwise such a constraint will be falsified). For instance, the least plausible interpretations are those which do not verify any $\Delta$-rule (namely, are not in any $A_k \cap B_k$).

**Algorithm 1:**
Data: $G$
Result: $WOP(\pi) = E_1 \cup \cdots \cup E_n$
**begin**
$\quad i \leftarrow 0;$
$\quad$**while** $(U \neq \emptyset)$ **do**
$\quad\quad i \leftarrow i + 1;$
$\quad\quad S_i = \{u : \not\exists \Delta(A_k \cap B_k) > \Delta(A_k \cap B_k^c)$ s.t. $u \in A_k \cap B_k\};$
$\quad\quad$**if** $S_i = \emptyset$ **then**
$\quad\quad\quad\lfloor$ $G$ is inconsistent
$\quad\quad$ - Remove from $U$ elements of $S_i$;
$\quad\quad$ - Remove from $\mathcal{C}$ constraints such that $S_i \cap (A_k \cap B_k^c) \neq \emptyset$
$\quad$**return** $E_1, \cdots, E_i$ s.t. $\forall j \leq i, E_j = S_{i-j+1}$
**end**

**Example 1** *Let us consider an example where* $G = \{p \leadsto q, p \leadsto \neg r, q \leadsto r\}$, *and* $p, q, r$ *are three propositional symbols. Let* $U = \{\omega_0 = pqr, \omega_1 = pq\neg r, \omega_2 = p\neg qr, \omega_3 = p\neg q\neg r, \omega_4 = \neg pqr, \omega_5 = \neg pq\neg r, \omega_6 = \neg p\neg qr, \omega_7 = \neg p\neg q\neg r\}.$
*Let us apply our previous algorithm. We have:*
$\mathcal{C} = \Delta(\{\omega_0, \omega_1\}) > \Delta(\{\omega_2, \omega_3\}), \Delta(\{\omega_1, \omega_3\}) > \Delta(\{\omega_0, \omega_2\}), \Delta(\{\omega_0, \omega_4\}) > \Delta(\{\omega_1, \omega_5\})\}.$
*The use of maximal specificity principle leads to first put* $\omega_6, \omega_7, \omega_2$ *and* $\omega_5$ *in the lowest rank since they are not constrained, namely from our algorithm, we have:*
$S_1 = \{\omega_6, \omega_7, \omega_2, \omega_5\}.$
*Once* $S_1$ *is settled, all the constraints in* $\mathcal{C}$ *are satisfied, hence* $S_2$ *contains the remaining interpretations.*
*Hence, the partition associated to* $G$ *is:*
$E_1 = \{\omega_0, \omega_1, \omega_3, \omega_4\}$ *and* $E_2 = \{\omega_2, \omega_5, \omega_6, \omega_7\}.$

A syntactic counterpart of the above algorithm where $G$ would be directly encoded in terms of $\Delta$-based formulas, can be provided by taking advantages of transformations between the different representations of a possibility distribution [2]. Its computational complexity is not more costly than the one used to transform a strict comparative possibility base into a possibilistic logic base [4].

## 6 Possibilistic networks

Another compact representation of a possibility distribution is graphical and is based on conditioning. Symbolic knowledge is represented by DAGs, where nodes represent variables (in this paper, we assume that they are binary), and edges express influence links between variables. When there exists a link from $X$ to $Y$, $X$ is said to be a parent of $Y$. The set of parents of a given node $X$ is denoted by $Par(X)$. By the capital letter $X$ we denote a variable which represents either the symbol $x$ or its negation. An interpretation in this section will be simply denoted by $X_1 \cdots X_n$. Uncertainty is expressed at each node by a pair of lower and upper possibility distributions as follows:
- For root nodes $X_i$ we provide the prior possibility degrees and guaranteed possibility degrees $(\Delta(X_i), \Pi(X_i))$ of $x_i$ and of its negation $\neg x_i$. These prior should satisfy the normalization condition: $max(\Pi(x_i), \Pi(\neg x_i)) = 1$. The normalization condition $min(\Delta(x_i), \Delta(\neg x_i)) = 0$ is assumed for guaranteed possibility degrees.
- For other nodes $X_j$, we provide $(\Delta(X_i \mid Par(X_i)), \Pi(X_i \mid Par(X_i)))$ i.e. the conditional possibility degrees and guaranteed possibility degrees of $x_j$ and of its negation $\neg x_j$ given any complete instantiation of each variable of parents of $X_j$, $u_{Par(X_j)}$. The conditional possibilities should also satisfy the normalization conditions:
$$max(\Pi(x_j \mid u_{Par(X_j)}), \Pi(\neg x_j \mid u_{Par(X_j)})) = 1.$$
$$min(\Delta(x_j \mid u_{Par(X_j)}), \Delta(\neg x_j \mid u_{Par(X_j)})) = 0.$$
The local possibilities $(\Delta(X_i), \Pi(X_i))$ (resp. $(\Delta(X_i \mid Par(X_i)), \Pi(X_i \mid Par(X_i)))$ should satisfy the coherence conditions:
$$\Delta(X_i) \leq \Pi(X_i)$$
(resp. $\Delta(X_i \mid Par(X_i)) \leq \Pi(X_i \mid Par(X_i))$).

A first approach to defining a pair of joint distributions associated with a bipolar possibilistic graph is to apply the following rules:

$$\pi^*(u) = min\{\Pi(x \mid u_{Par(X)}) : u \models x \text{ and } u \models u_{Par(X)}\}, \quad (13)$$

$$\pi_*(u) = max\{\Delta(x \mid u_{Par(X)}) : u \models x \text{ and } u \models u_{Par(X)}\}. \quad (14)$$

**Example 2** *Let us consider the following graph.*

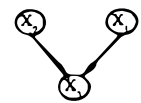

Figure 1:

*We have:* $(\Delta(x_1) = 0, \Pi(x_1) = .5),$
$(\Delta(x_2) = 0, \Pi(x_2) = 1), (\Delta(\neg x_1) = .45, \Pi(\neg x_1) = 1),$
$(\Delta(\neg x_2) = .4, \Pi(\neg x_2) = .5).$
*Table 1 gives the conditional possibilities* $(\Delta(X_3 \mid X_1 X_2), \Pi(X_3 \mid X_1 X_2)).$
*Table 2 shows the possibility distributions* $\pi^*$ *and* $\pi_*$ *associated with the graph.*

|  | $x_1 x_2$ | $x_1 \neg x_2$ | $\neg x_1 x_2$ | $\neg x_1 \neg x_2$ |
|---|---|---|---|---|
| $x_3$ | $(.4, 1)$ | $(0, 1)$ | $(0, .5)$ | $(.4, .8)$ |
| $\neg x_3$ | $(0, .6)$ | $(.3, 1)$ | $(.2, 1)$ | $(0, 1)$ |

Table 1:

| $u$ | $\pi^*(u)$ | $\pi_*(u)$ |
|---|---|---|
| $\neg x_1 \neg x_2 \neg x_3$ | .5 | .45 |
| $\neg x_1 \neg x_2 x_3$ | .5 | .45 |
| $\neg x_1 x_2 \neg x_3$ | 1 | .45 |
| $\neg x_1 x_2 x_3$ | .5 | .45 |
| $x_1 \neg x_2 \neg x_3$ | .5 | .4 |
| $x_1 \neg x_2 x_3$ | .5 | .4 |
| $x_1 x_2 \neg x_3$ | .5 | 0 |
| $x_1 x_2 x_3$ | .5 | .4 |

Table 2:

It is easy to check that $\forall u, \pi_*(u) \leq \pi^*(u)$.
However, the coherence property is not always guaranteed. Indeed, consider a bipolar possibilistic graph composed of two distinct nodes $X$ and $Y$, such that:
$(\Delta(x), \Pi(x)) = (.3, .3), (\Delta(y), \Pi(y)) = (.5, .6)$
$(\Delta(\neg x), \Pi(\neg x)) = (0, 1), (\Delta(\neg y), \Pi(\neg y)) = (0, 1)$.
We have $\pi^*(xy) = .3$ while $\pi_*(xy) = .5$.
One way to guarantee that $\pi_* \leq \pi^*$ is to replace (14) by:

$$\pi_*(u) = min(min\{\Pi(x \mid u_{Par(X)}) : u \models x \text{ and } u \models u_{Par(X)}\},$$
$$max\{\Delta(x \mid u_{Par(X)}) : u \models x \text{ and } u \models u_{Par(X)}\}).$$

This expresses a revision of the lower distribution by means of the upper distributions in order to restore consistency.

## 7 Diagnosis

Diagnosis knowledge can be described by means of relations between causes and effects [14]. The effects of a cause $c$ usually pertain to several attributes describing the system under consideration. On a given attribute, the possible effects of $c$ are usually represented by a *subset* of possible values rather than a single value.

For instance, we may have the rule "if you have a flu, your fever is in the interval $[38.5, 40]^\circ C$". If we assume that our information is *complete* on the possible effects of a flu on the fever, it means that both *i)* any fever outside $[38.5, 40]^\circ C$ is impossible, and *ii)* any fever in $[38.5, 40]^\circ C$ is feasible, if you have a flu.

Suppose our knowledge about possible effects is uncertain rather than complete. Then, we have to use a distribution on the *power set* of the attribute domain. In the possibilistic framework this distribution will encode that some subsets are more plausible than others for describing the exact subset of possible values for the attribute when the cause takes place. Possibility distributions on a power set are heavy to handle and can be then approximated by a consistent pair of distributions on the domain we start with [7]. In practice, this means that a lower distribution describes the set of values which are possible effects more or less certainly, i.e. which are guaranteed as being possible effects, while the upper distribution restricts the values which are not impossible to some extent as being possible effects.

Let $\pi_{c*}^i$ and $\pi_c^{i*}$ be the lower and upper distributions representing what is known of the effects of $c$ on attribute $i$. Given an observation of the value of attribute $i$, which may be pervaded with imprecision and uncertainty, represented by a fuzzy set $O_i$, two indices can be computed for rank-ordering the causes according to their likelihood of presence. Namely, the consistency index of $c$ with the observation defined by

$$cons(O_i, \pi_c^{i*}) = sup_u \, min(O_i(u), \pi_c^{i*}(u)) \quad (15)$$

evaluates to what extent there exists a value compatible with the observation which is not impossible for $i$ when $c$ takes place.

Another index of relevance of $c$ w.r.t. the observation can be defined as

$$rel(O_i, \pi_{c*}^i) = inf_u \, O_i(u) \Rightarrow \pi_{c*}^i(u) \quad (16)$$

where $\Rightarrow$ denotes an implication function. This estimates to what extent all the possible values of $i$ compatible with the observation are among the values which are guaranteed to be possible effects when $c$ is present. When $O_i$ is an ordinary subset of the domain of attribute $i$, we can easily recognize that $cons(O_i, \pi_c^{i*})$ is nothing $\Pi_c(O_i) = sup_{u \in O_i} \pi_c^{i*}(u)$ while $rel(O_i, \pi_{c*}^i)$ is nothing but $\Delta_c(O_i) = inf_{u \in O_i} \pi_{c*}^i(u)$. Thus, when observation is precise, (15) expresses that $O_i$ is among the effects of $c$ which are not impossible, while (16) indicates if $O_i$ is among the guaranteed possible effects of $c$. (15) and (16) extend $\Pi_c$ and $\Delta_c$ to fuzzy sets. It can be shown that using Gödel implication in (16), namely

$x \Rightarrow y = 1$ if $x \leq y$ and $x \Rightarrow y = y$ if $x > y$.
preserves the equivalence $\Delta(A) \geq \alpha \Leftrightarrow \pi_*(u) \geq min(A(u), \alpha)$ for fuzzy sets $A$.
Thus, clearly the relevance index is such that $\Delta_c(O_i) \leq \Pi_c(O_i)$, and thus refines the first one.

Besides, from (11), we can write the Bayesian-like equation
$Mm((\Delta, \Pi)(A \mid B), (\Delta, \Pi)(B)) = Mm((\Delta, \Pi)(B \mid A), (\Delta, \Pi)(A)).$
In case of empty prior $(\Delta, \Pi)(B) = (0, 1)$, and using a postulate of "observation relevance" [9] stating that
$mM((\Delta, \Pi)(A \mid B), (\Delta, \Pi)(A \mid B^c)) = (0, 1),$
it can be shown that $(\Delta, \Pi)(B \mid A)$ behaves like $(\Delta, \Pi)(A \mid B)$. This provides a justification for using

$(\Delta_c, \Pi_c)(O_i)$ for estimating the plausibility of having cause $c$ given $O_i$.

## 8 Conclusion

This paper has presented and advocated a twofold framework for representing uncertainty qualitatively. This can be useful for knowledge representation as well as preference representation. The bipolarity enables us to distinguish between what is possible or satisfactory for sure from what is just not impossible, or not undesirable.

It has been recently shown that bipolar possibilistic logic provides a natural framework for representing positive and negative desires [3]. Indeed, negative desires which refer to what is more or less rejected by an agent, correspond to constraints of the form $\Pi(R_k) \leq \rho_k$. They express that the models of $R_k$ are considered as rather impossible. Positive desires are represented by constraints of the form $\Delta(D_j) \geq \delta_j$, which expresses that any model of $d_j$ is satisfactory at least at level $\delta_j$. This representation framework has been applied in [3] for fusing the positive and negative preferences of different agents. Even if the preferences of each agent $t$ are consistent in the sense that $\forall u, \pi_*^t \leq \pi^{t*}$, it may occur, that the result of the fusion of the lower distributions is no longer consistent with the result of the fusion of the upper distributions, indeed the same aggregation operator is not used for fusing the positive preferences represented by $\pi_*^t$ on the one hand, and the non-rejection-based preferences encoded by $\pi^{t*}$ on the other hand. Then a revision step should take place for enforcing the consistency.

Besides, a possibility distribution can be represented equivalently in various compact frameworks: as a possibilistic logic base, as a possibilistic graph, or as strict comparative possibility constraints. This has been extended to the bipolar setting, by first defining the idea of conditioning (underlying the two last compact representation frameworks listed above) for guaranteed possibility measures. Algorithms exist for going directly from one compact representation to another [2]. They can be adapted to the bipolar framework. Efficient algorithms recently developed for uncertainty propagation in possibilistic graphs [1] could be extended to bipolar possibilistic graphs as well.

Lastly, the bilattice structure [11] of the bipolar framework is worth investigating and may be of interest, especially in the view of developing possibilistic logic programming, following Fitting [10]'s approach.